\documentclass[letterpaper, 10 pt, conference]{ieeeconf}  

\IEEEoverridecommandlockouts                              

\usepackage{graphicx}
\usepackage{amsmath}
\usepackage{array}
\usepackage{booktabs}
\usepackage{multirow}
\usepackage{xcolor}
\usepackage{soul}
\usepackage{hyperref}
\usepackage[normalem]{ulem}
\usepackage[font=footnotesize,labelformat=simple]{subcaption}

\soulregister{\cite}{7}
\sethlcolor{yellow}

\title{\LARGE \bf
MOVE: Multi-skill Omnidirectional Legged Locomotion with Limited View in 3D Environments
}

\author{Songbo Li$^{\dagger1}$ Shixin Luo$^{\dagger1}$ Jun Wu$^{1,2}$ Qiuguo Zhu$^{*1,2}$ 
\thanks{This work was supported by the National Key R\&D Program of China (Grant No. 2022YFB4701502), the ”Leading Goose” R\&D Program of Zhejiang (Grant No. 2023C01177), and the 2035 Key Technological Innovation Program of Ningbo City (Grant No. 2024Z300).}
\thanks{The supplementary video is available at \url{https://youtu.be/Z3P2voPS_F0}.}
\thanks{$^{1}$The authors are with Institute of Cyber-Systems and Control, Zhejiang University, 310027, China.}%
\thanks{$^{2}$Qiuguo Zhu and Jun Wu are with State Key Laboratory of Industrial Control Technology, 310027, China.}%
\thanks{$^\dagger$These authors contributed equally to this work.}%
\thanks{$^*$ Qiuguo Zhu ({\tt\small qgzhu@zju.edu.cn}) is the corresponding author.}%
}

\begin{document}

\maketitle
\thispagestyle{empty}
\pagestyle{empty}

\begin{abstract}

Legged robots possess inherent advantages in traversing complex 3D terrains. However, previous work on low-cost quadruped robots with egocentric vision systems has been limited by a narrow front-facing view and exteroceptive noise, restricting omnidirectional mobility in such environments. While building a voxel map through a hierarchical structure can refine exteroception processing, it introduces significant computational overhead, noise, and delays. In this paper, we present MOVE, a one-stage end-to-end learning framework capable of multi-skill omnidirectional legged locomotion with limited view in 3D environments, just like what a real animal can do. When movement aligns with the robot's line of sight, exteroceptive perception enhances locomotion, enabling extreme climbing and leaping. When vision is obstructed or the direction of movement lies outside the robot's field of view, the robot relies on proprioception for tasks like crawling and climbing stairs. We integrate all these skills into a single neural network by introducing a pseudo-siamese network structure combining supervised and contrastive learning which helps the robot infer its surroundings beyond its field of view. Experiments in both simulations and real-world scenarios demonstrate the robustness of our method, broadening the operational environments for robotics with egocentric vision.
\end{abstract}

\section{INTRODUCTION}

Legged robots have garnered significant attention due to their advanced locomotion capabilities. With a body structure analogous to that of animals, legged robots can flexibly adjust their footholds and postures to navigate through complex environments. In recent years, advancements in learning-based methods have continuously unlocked the potential of legged robots, enabling them to traverse uneven terrain, avoid obstacles, walk through confined spaces, and even perform highly dynamic motions~\cite{hwangbo2019learning,miki2022learning,margolis2023walk,fu2022coupling,miki2024learning,zhang2023learning,luo2024pie}.

Due to inaccuracy and instability introduced by commonly used exteroceptive sensors, such as depth cameras and LiDARs, recent studies have proposed various proprioception-only methods to avoid these issues~\cite{lee2020learning,kumar2021rma,nahrendra2023dreamwaq,long2023hybrid,cheng2024quadruped,wang2024combining}. However, robots still require vision to avoid hazardous trial-and-error processes and traverse more complex terrains. As a result, some researchers have developed decoupled pipelines that separate terrain perception from motion control~\cite{miki2022learning,hoeller2024anymal,miki2024learning}. In these systems, the upper-level perception system provides the controller with a relatively reliable elevation or voxel map~\cite{miki2022elevation,hoeller2022neural}. Nevertheless, this approach presents additional challenges, including increased noise, redundant information and unsuitability for low-cost robots, as it typically requires multiple cameras or LiDAR sensors for real-time mapping. Moreover, the mapping process usually relies on smooth exteroception and is time-consuming, making it difficult to apply in dynamic scenarios.

\begin{figure}[t]
\centering
\includegraphics[width=\columnwidth]{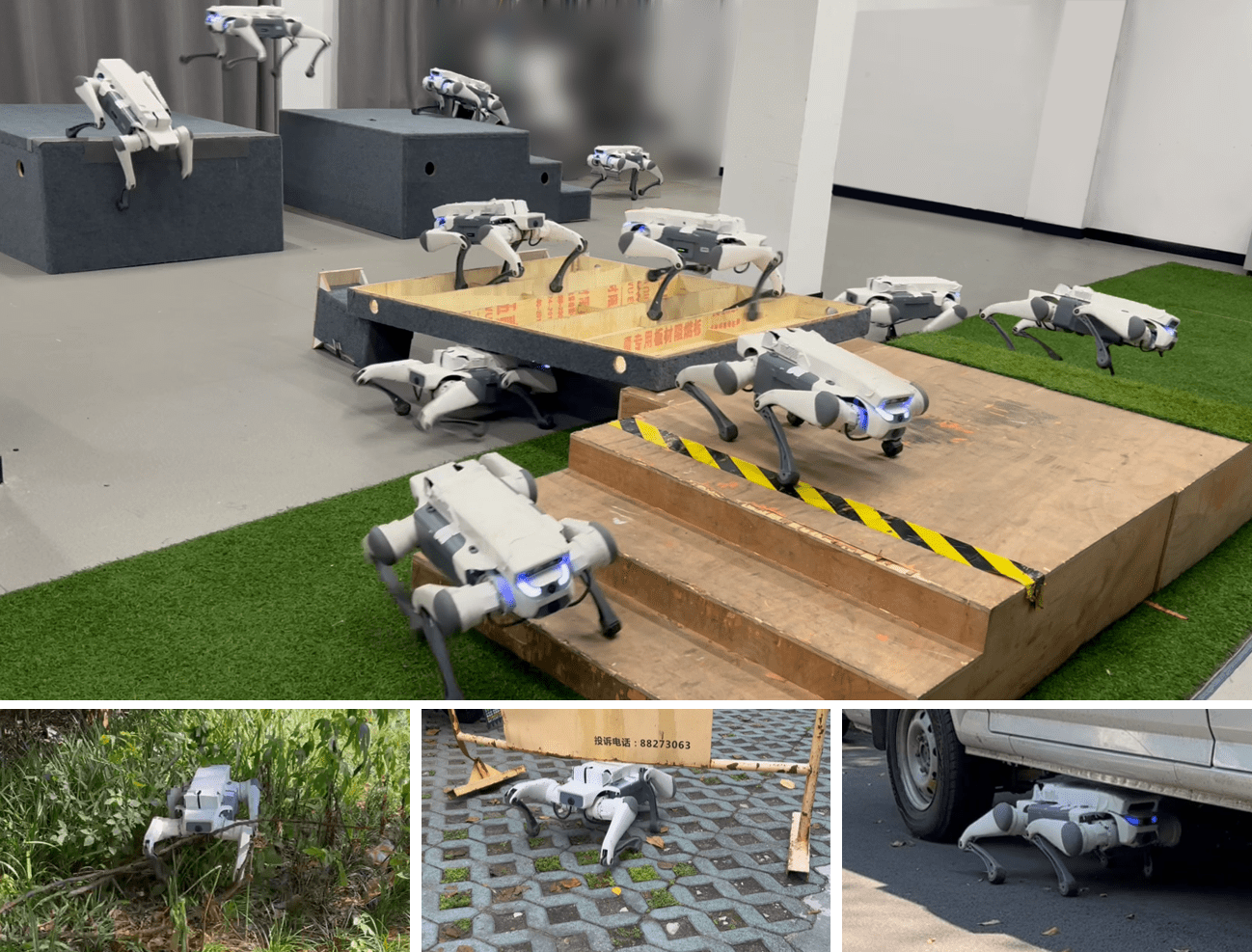}
\vspace{-0.5cm}
\captionsetup{font=footnotesize}
\caption{We deploy our policy in real-world environments, demonstrating the diverse and exceptional motion skills of the proposed framework MOVE. Our robot utilizes limited visual perception, successfully traversing complex 3D environments omnidirectionally. Even when the exteroception was severely disrupted or unavailable, the vision-dependent policy still enables the robot to overcome obstacles.}
\vspace{-0.7cm}
\label{fig_1}
\end{figure}

Recent research in training end-to-end vision-aided policies for legged robots has achieved significant success to potentially address issues associated with hierarchical learning framework~\cite{yang2021learning,agarwal2023legged,loquercio2023learning,yang2023neural,zhuang2023robot,cheng2023extreme,luo2024pie}. However, developing a comprehensive motion control strategy for current mainstream low-cost legged robots with egocentric vision still faces several challenges. Chief among these is the problem of incomplete visual perception. Current research primarily focuses on enabling robots to overcome challenging terrains within their front-facing visual range. These approaches work reasonably well for forward locomotion, where most necessary information can be obtained through vision and short-term memory. However, in omnidirectional motion tasks within complex environments, such unidirectional visual observation may cause the network to fail in extracting the information needed for effective movement.

Despite relying on limited egocentric vision, animals can tightly integrate their visual and proprioceptive senses, enabling them to infer terrain properties necessary for locomotion even beyond their line of sight. For instance, when an animal unexpectedly trips or bumps its head against an obstacle, it can quickly adjust and recover from the unseen impact. Inspired by this, it is crucial for legged robots to learn the intrinsic relationships between seen and unseen surroundings. When the direction of movement aligns with the line of sight, exteroceptive perception should be fully utilized to enhance locomotion. Conversely, when vision is obstructed or the direction of movement is outside the field of view, the robot should rely on proprioception for inference. All these capabilities should be seamlessly integrated into a single network, eliminating the need for manual switching.

In this paper, we propose a one-stage end-to-end learning-based framework MOVE, which is capable of multi-skill omnidirectional legged locomotion with limited
view in 3D environments and addresses all the aforementioned challenges. We employ a combination of contrastive learning and reconstruction approach for representation learning, aiming to extract effective features from incomplete and noisy inputs. Our results on a low-cost quadruped robot with an egocentric depth camera demonstrate that the framework enables the robot to infer environmental conditions and execute dynamic omnidirectional movements in complex environments, including traversing 3D obstacles and performing strenuous jumps, even under incomplete and noisy exteroceptive conditions.

Our contributions are as follows:
\begin{itemize}
\item{A one-stage end-to-end learning framework that combines contrastive learning and reconstruction approach is proposed to inspire omnidirectional traversability on a variety of challenging terrains under limited visual perception.}
\item{A pseudo-siamese representation learning method via asymmetric attention mechanism is introduced to extract terrain features. This method enables the robot not only to interpret the environment within field of view but also to infer surroundings where it can’t directly “see”.}
\item{Extensive real-world experiments are conducted in diverse and complex environments to validate the effectiveness and the robustness of the proposed framework.}
\end{itemize}

\section{RELATED WORK}
\label{chap:2}

\subsection{Blind Locomotion}

Classical model-based techniques have achieved some promising results in legged locomotion~\cite{miura1984dynamic,raibert1984hopping,geyer2003positive,yin2007simbicon,sreenath2011compliant,ames2014rapidly,bledt2018cheetah,hutter2016anymal}, enabling legged robots to achieve fast walking and jumping. However, these methods often struggle to adaptively select dynamic and agile behaviors in complex situations. Inspired by the way animals learn robust locomotion skills, some reinforcement learning (RL) methods have been proposed and show considerable promise. Recent learning-based approaches primarily focus on utilizing proprioceptive sensors to gain insights into both the robot itself and its surrounding environment. This is achieved by fitting responses to privileged information~\cite{lee2020learning,kumar2021rma,fu2023deep,wang2024combining}, estimating the robot's state and its transition processes~\cite{ji2022concurrent,nahrendra2023dreamwaq,long2023hybrid,luo2024moral,wang2024combining}, performing collision/contact detection~\cite{lin2021legged,cheng2024quadruped}, or incorporating imitation learning~\cite{wu2023learning}. However, policies trained solely with proprioceptive feedback often struggle to handle highly challenging terrains.

\begin{figure*}[t]
\centering
\vspace{0.2cm}
\includegraphics[width=6in]{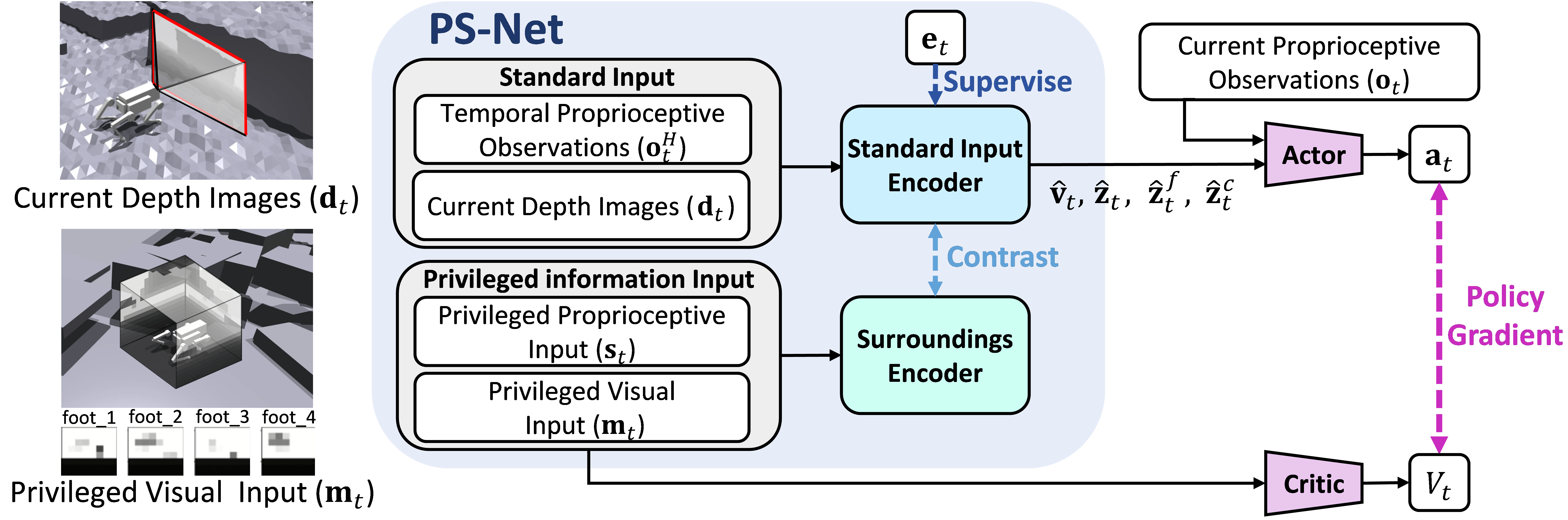}
\vspace{-0.2cm}
\captionsetup{font=footnotesize}
\caption{Overview of the proposed MOVE framework. We use a one-stage learning pipeline to train a comprehensive locomotion policy with access to limited visual perception. The left sides of the figure illustrates the composition of two types of the visual input. The blue box represents PS-Net, which is trained by a combination of supervised and unsupervised learning method.}
\vspace{-0.7cm}
\label{fig_2}
\end{figure*}

\subsection{Vision-aided Locomotion}

Incorporating visual perception significantly enhances the locomotion performance of legged robots, allowing them to flexibly select appropriate behaviors based on their anticipation of complex situations. Inspired by how animals perceive the world, egocentric vision has recently been employed in learning frameworks, yielding substantial achievements~\cite{yang2021learning,agarwal2023legged,loquercio2023learning,yang2023neural,zhuang2023robot,cheng2023extreme,luo2024pie,han2024lifelike}. However, most of them focuses on forward motion and assumes undisturbed visual input rather than developing a comprehensive locomotion strategy. Other methods use elevation or voxel maps as visual inputs to achieve robust locomotion in complex and even confined spaces~\cite{miki2022learning,rudin2022advanced,hoeller2024anymal,miki2024learning}. Nevertheless, real-time mapping may fail on low-cost robots with limited onboard computing and sensors, and is prone to significant noise and drift when the robot performs highly dynamic maneuvers.

\section{METHOD}
\label{chap:3}

\subsection{Overview}

\subsubsection{Overall Structure}

Our framework consists of four main components: a standard input encoder, a surroundings encoder, a policy network, and a value network. The standard input encoder processes standard input that can be directly obtained from the real robot, outputting a latent vector that serves as the extracted features for the actor to generate actions. During deployment, only this pathway is used. The surroundings encoder utilizes inputs obtainable only in simulation, providing a more comprehensive encoding of the surrounding environment. This encoder, along with the reconstruction of ground truth estimates $\mathbf{e}_t$, ensures the latent vector output by the standard input encoder has sufficient understanding of its surroundings to guide the action decision process. The value network also receives privileged information, and the PPO algorithm is used to optimize the asymmetric actor-critic framework~\cite{schulman2017proximal}.

\subsubsection{Composition of Network Input}

The standard input comprises two components: the observation history $\mathbf{o}_t^H$ and the depth image $\mathbf{d}_t$. The proprioception is defined as
\begin{equation}
  \mathbf{o}_t=\begin{bmatrix}\boldsymbol{\omega}_{t} & \mathbf{g}_{t} & \mathbf{v}_{t}^{cmd} & {\omega}_{t}^{cmd} & \boldsymbol{\theta}_{t} & \dot{\boldsymbol{\theta}}_{t} & \mathbf{a}_{t-1} \end{bmatrix}^T,
\end{equation}
which includes the body angular velocity $\boldsymbol{\omega}_{t}$, the gravity direction vector in the body frame $\mathbf{g}_{t}$, the linear velocity commands in the $xy$ plane $\mathbf{v}_{t}^{cmd}$, the yaw rate command ${\omega}_{t}^{cmd}$, the joint angle $\boldsymbol{\theta}_{t}$, the joint angular velocity $\dot{\boldsymbol{\theta}}_{t}$ and the previous action $\mathbf{a}_{t-1}$. To retain transient memory, we stack the proprioception over the past $H$ steps. In this work, $H$ is set to 10. The depth image $\mathbf{d}_t$ is obtained from a single frame captured by a front-facing depth camera.

The privileged information input also comprises proprioceptive and exteroceptive components. The proprioceptive component includes proprioception and linear velocity, which is defined as
\begin{equation}
\mathbf{s}_t=\begin{bmatrix}\mathbf{o}_{t} & \mathbf{v}_{t} \end{bmatrix}^T.
\end{equation}
In the context of 3D complex environment locomotion tasks, it is challenging to fully describe terrain properties using an elevation map alone. Given that a 2.5D elevation map is incomplete and a 3D voxel map is computationally and storage-intensive, we propose a novel privileged visual observation $\mathbf{m}_t$, which provides a lightweight and efficient 3D terrain representation, defined as 
\begin{equation}
\mathbf{c}_{t}=\begin{bmatrix}\mathbf{c}_{t}^{front} & \mathbf{c}_{t}^{upward} & \mathbf{c}_{t}^{downward} & \mathbf{c}_{t}^{left} & \mathbf{c}_{t}^{right} \end{bmatrix}^T,
\end{equation}
\begin{equation}
\mathbf{m}_t=\begin{bmatrix}\mathbf{c}_{t} & \mathbf{l}_{t}^{foot} \end{bmatrix}^T.
\end{equation}
Here, $\mathbf{c}_{t}$ represents a cube map composed of five egocentric depth images oriented forward, upward, downward, left, and right, excluding the rear, as the robot's front-facing perception and short-term memory are insufficient to establish a meaningful connection with rearward data. This cube map sampling method captures depth information more uniformly across space compared to equidistant rectangular projections obtained through LiDAR-like sampling, which tend to oversample at the poles and undersample at the equator. Additionally, $\mathbf{l}_t^{foot}$ contains sparse depth information around the robot's four feet.

\begin{figure*}[t]
\centering
\vspace{0.2cm}
\includegraphics[width=5.5in]{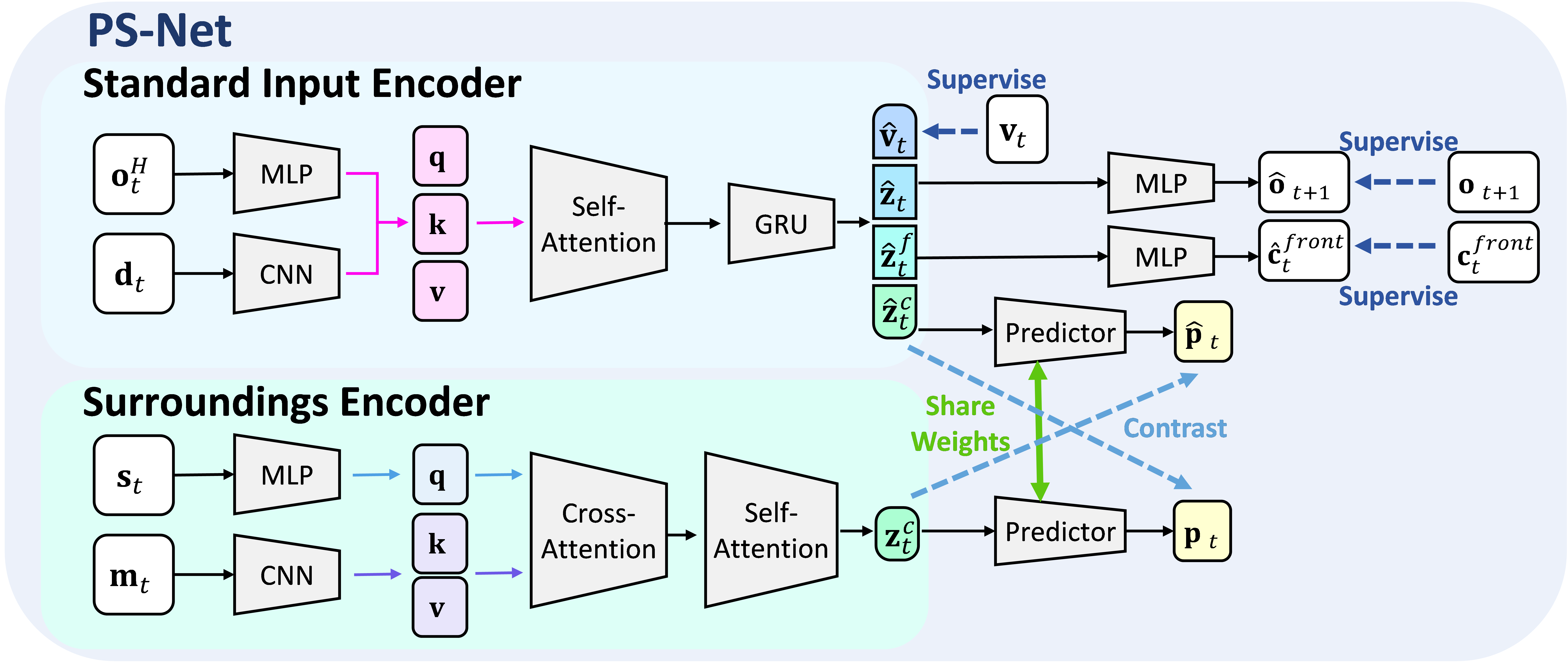}
\vspace{-0.1cm}
\captionsetup{font=footnotesize}
\caption{The detailed architecture of PS-Net, consisting two main parts: standard input encoder and surroundings encoder. Utilizing a pseudo-siamese network structure, PS-Net is able to extract shared feature between standard input and priviledged information even from incomplete and noisy observations. }
\vspace{-0.5cm}
\label{fig_3}
\end{figure*}

\subsubsection{Policy Network and Value Network}

The policy network takes proprioception $\mathbf{o}_t$ and the latent vector output by the standard input encoder as inputs to decide a 12-dimensional action vector $\mathbf{a}_t$.

The value network first encodes the privileged information using a critic encoder, which involves applying an MLP and CNN to project $\mathbf{s}_t$ and $\mathbf{m}_t$, respectively, and then employing a transformer to fuse the multi-modal tokens. The state value $V_t$ is subsequently computed using an MLP.

\subsubsection{Reward Function}

Following previous work~\cite{luo2024pie}, we adopt the same simple yet effective reward function with minimal modifications. We demonstrate that even without extensive parameter tuning, a policy learned with sufficient environmental understanding can significantly enhance the locomotion capabilities of legged robots.

\begin{figure}[t]
\centering
\includegraphics[width=0.8\columnwidth]{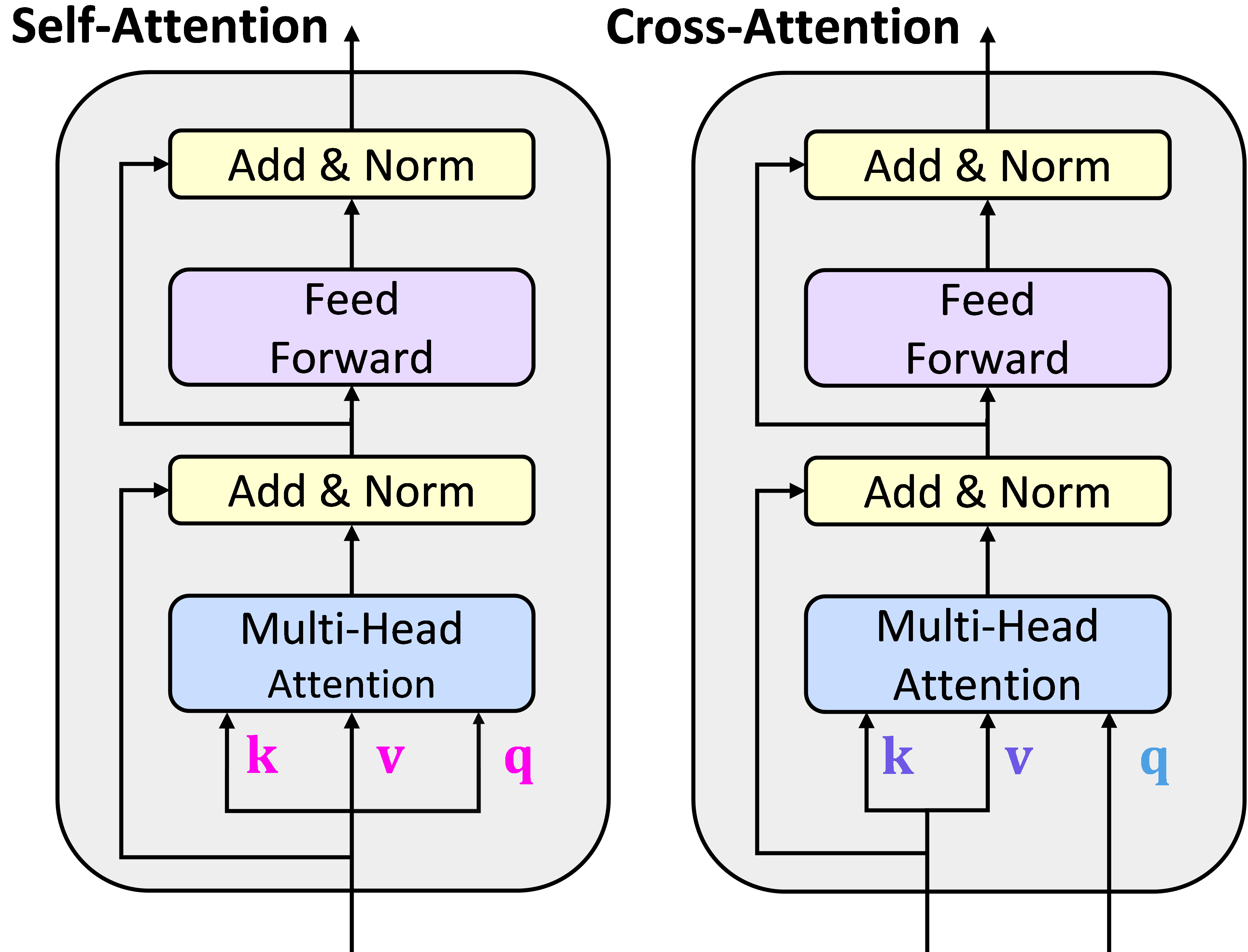}
\vspace{-0.1cm}
\captionsetup{font=footnotesize}
\caption{Asymmetric attention mechanism in PS-Net. PS-Net incorporates an asymmetric attention mechanism to effectively process different input modalities. The standard input encoder employs a self-attention mechanism to fuse multimodal information from proprioception and depth images. Conversely, the surroundings encoder utilizes a cross-attention module to focus on privileged visual inputs to help standard input encoder extract surrounding visual features by contrastive learning.}
\vspace{-0.7cm}
\label{fig_4}
\end{figure}

\subsection{PS-Net}
\label{chap:3.2}

While the reconstruction method effectively compresses input into latent representation with specific meanings, it faces challenges when the connection between input and required estimation is weak. To address this limitation, we introduce a pseudo-siamese network structure (PS-Net) with mixed representation learning approach. By using a similar, yet non-weight-sharing network structure, PS-Net can capture rich environmental features, allowing the robot to infer its surroundings from limited observations.

\subsubsection{Standard Input Encoder}

The standard input encoder uses an MLP to process the observation history $\mathbf{o}_t^H$ to obtain the proprioception features and a CNN to extract visual features from the depth image $\mathbf{d}_t$. These proprioceptive and exteroceptive features are then concatenated and processed by a transformer encoder using self-attention mechanism for improved cross-modality integration. To preserve the agent's long-term memory, the output of the transformer is fed into a GRU, ultimately producing the latent vector, which includes $\hat{\mathbf{v}}_t$, $\hat{\mathbf{z}}_t$, $\hat{\mathbf{z}}_t^f$ and $\hat{\mathbf{z}}_t^c$. Details on the latent vector will be elaborated upon later.

\subsubsection{Surroundings Encoder}

The surroundings encoder, similar to the value network, processes privileged information and employs a comparable network structure. However, we do not use a weight-sharing encoder for these two encoders as their encoding serves distinct purposes: the critic encoder needs to evaluate from a global perspective to provide value references, whereas the surroundings encoder aims to help standard input encoder extract surrounding visual features by contrastive learning. A key difference is the introduction of the cross-attention mechanism in the surroundings encoder to handle multimodal inputs. Specifically, the query input $\mathbf{q}$ is computed by projecting the privileged proprioceptive observation $\mathbf{s}_t$ through an MLP, while the key input $\mathbf{k}$ and value input $\mathbf{v}$ are computed by projecting the visual embedding, which is derived from the privileged visual observation $\mathbf{m}_t$ via CNN. In the subsequent Add \& Norm module, only the visual embedding participates in the skip connection. This mechanism is intended to ensure that the surroundings encoder focuses on its visual input, with the privileged proprioceptive observation serving as auxiliary information through the attention mechanism. Without this, the surroundings encoder struggles to prioritize visual information, leading to the collapse of the contrastive learning because of the highly similar proprioception $\mathbf{o}_t$ and $\mathbf{s}_t$, thereby losing some implicit environmental information.

\subsubsection{Training Process}

The training of PS-Net employs a mixed loss fuction: 
\begin{equation} 
\mathcal{L} = \mathcal{L}_{reconstruction} + \mathcal{L}_{contrast}. 
\end{equation}
Here, $\mathcal{L}_{reconstruction}$ and $\mathcal{L}_  {contrast}$ represent the loss functions introduced in this work for the reconstruction and contrastive learning methods, respectively.

The reconstruction method focuses on features that can be relatively easily extracted, utilizing an encoder-decoder structure to ensure that the extracted features encapsulate the information contained in the ground truth estimates $\mathbf{e}_t$. To be specific, the velocity estimate $\hat{\mathbf{v}}_t$ within the latent vector is directly regressed against the ground truth $\mathbf{v}_t$, and the implicit state estimation $\hat{\mathbf{z}}_t$, which employs a VAE structure, is reconstructed through an MLP decoder to $\hat{\mathbf{o}}_{t+1}$, and then regressed against the ground truth $\mathbf{o}_{t+1}$. To enhance the robot’s intuitive understanding of ${\mathbf{d}}_t$, another vision estimation $\hat{\mathbf{z}}_t^f$ is decoded by an MLP into the front-facing depth map $\hat{\mathbf{c}}_t^{front}$ from the cube map, and the regression error is computed. The reconstruction loss for the above is calculated using mean-square error (MSE):
\begin{equation} 
\begin{aligned}
\mathcal{L}_{reconstruction} = &D_\text{KL}(q(\hat{\mathbf{z}}_t|\mathbf{o}^{H}_{t}, \mathbf{d}_{t})\parallel p(\hat{\mathbf{z}}_{t}))\\
&+ \text{MSE}(\hat{\mathbf{o}}_{t+1}, \mathbf{o}_{t+1}) + \text{MSE}(\hat{\mathbf{v}}_{t}, \mathbf{v}_{t})\\
&+ \text{MSE}(\hat{\mathbf{c}}_{t}^{front}, \mathbf{c}_{t}^{front}).
\end{aligned}
\end{equation}

\begin{table*}[t]
    \footnotesize
    \centering
    \vspace{0.2cm}
    \captionsetup{font=footnotesize, singlelinecheck=false}
    \caption{Ablation study for the representation learning methods. We tested each policy with success rates of various motion skills in simulation.}
    \label{table:1}
    
    \begin{center}
    \vspace{-0.3cm}
    \scriptsize
    \begin{tabular}{>{\centering\arraybackslash}m{2cm}>{\centering\arraybackslash}m{2cm}*{5}{>{\centering\arraybackslash}m{1.9cm}}}
        \toprule
        \multirow{2}{*}{Skills} & \multirow{2}{*}{Commands} &
        \multicolumn{5}{c}{Success Rate} \\  
        \cmidrule(lr){3-7} & 
        & Ours & Ours w/o C.L & Ours w/o Recon & Ours w/o C.A & Baseline \\ 
        \midrule
        High Jump & Forward & \textbf{99.7} & 81.2 & 48.2 & 96.5 & 31.1 \\
        \midrule
        Long Jump & Forward & \textbf{99.6} & 50.6 & 49.0 & 91.2 & 7.2 \\
        \midrule
        \multirow{3}{*}{Stairs Traversal} & Forward & \textbf{96.3} & 84.5 & 75.4 & 93.4 & 47.3 \\
        & Lateral & \textbf{97.1} & 88.5 & 80.8 & 96.2 & 36.2 \\
        & Backward & \textbf{99.2} & 98.2 & 98.2 & 98.8 & 61.4 \\
        \midrule
        \multirow{3}{*}{Crawl} & Forward & \textbf{91.8} & 81.7 & 26.5 & 86.6 & 71.8 \\
        & Lateral & \textbf{84.9} & 80.2 & 79.8 & 81.4 & 73.1 \\
        & Backward & \textbf{89.3} & 83.1 & 88.6 & 87.3 & 74.8 \\
        \midrule
        Blind Crawl & Omnidirectional & \textbf{80.7} & 75.3 & 65.1 & 71.6 & 66.3 \\
        \midrule
        Camera-offset Traversal & Forward & \textbf{98.0} & 94.6 & 94.1 & \textbf{98.0} & 76.5 \\
        
        \bottomrule
    \end{tabular}
    \end{center}
    \vspace{-0.2cm}
\end{table*}

In the contrastive learning component, the latent vector component $\hat{\mathbf{z}}_t^c$ and the output of the surroundings encoder $\mathbf{z}_t^c$ should capture the shared features between the standard input and the privileged information input. To achieve this, $\hat{\mathbf{z}}_t^c$ and $\mathbf{z}_t^c$ are transformed by a weight-sharing predictor, producing alternative views of the encoded features. Denoting the two output vectors as $\hat{\mathbf{p}}_t^c$ and $\mathbf{p}_t^c$, we minimize their negative cosine similarity as
\begin{equation} 
\begin{aligned}
\mathcal{L}_{constrast} = &-\frac{\mathbf{p}_t^c}{||\mathbf{p}_t^c||_{2}}\cdot\frac{\text{stopgrad}(\hat{\mathbf{z}}_t^c)}{||\text{stopgrad}(\hat{\mathbf{z}}_t^c)||_{2}}\\
&-\frac{\hat{\mathbf{p}}_t^c}{||\hat{\mathbf{p}}_t^c||_{2}}\cdot\frac{\text{stopgrad}(\mathbf{z}_t^c)}{||\text{stopgrad}(\mathbf{z}_t^c)||_{2}}.
\end{aligned}
\end{equation}
The stop-gradient ($\text{stopgrad}$) operation effectively prevents collapsing solutions, which occur when the model outputs converge to a constant value in pursuit of representational similarity~\cite{chen2021exploring}.

\section{EXPERIMENTS}
\label{chap:4}

\subsection{Experimental Setup}

We used Isaac Gym simulator to train the entire network in a single stage. To develop a comprehensive policy, we designed a series of simulation environments that cover the learning of skills such as leaping, crawling, climbing stairs, and omnidirectional walking on discrete terrain. Moreover, we introduced patterned noise into the depth image during simulation to mimic the noise that may arise when occlusions occur in real-world scenarios. In each environment, we incorporated a terrain curriculum with increasing difficulty~\cite{rudin2022learning}.

We utilized a DEEP Robotics Lite3 to assess the multi-skill locomotion capabilities of the proposed framework. Lite3 is a low-cost quadruped robot equipped with a front-facing Intel RealSense D435i. When controlled by its built-in MPC controller, it has a bounding box of 610mm $\times$ 37mm $\times$ 445mm~\cite{deeprobotics2024lite3}. For deployment, all computations of inference were processed by an onboard RK3588.

\begin{figure}[t]
\centering
\vspace{-0.3cm}
\includegraphics[width=0.8\columnwidth]{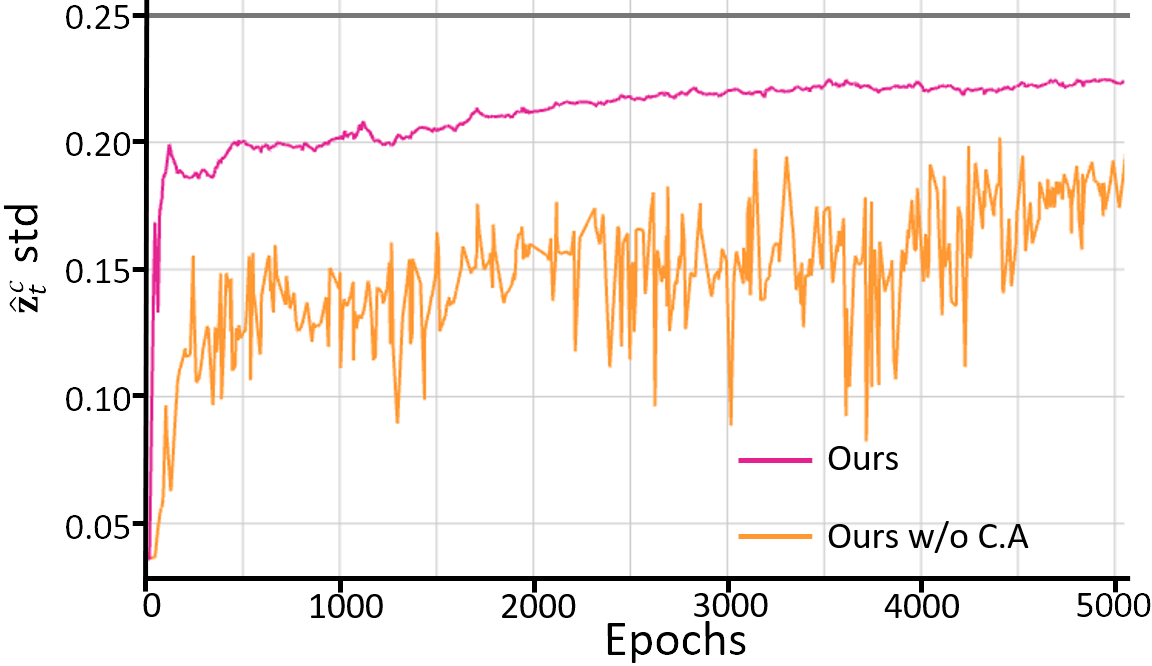}
\vspace{-0.1cm}
\captionsetup{font=footnotesize}
\caption{The average per-channel std of the ${\ell}_2$-normalized $\hat{\mathbf{z}}_t^c$. If $\hat{\mathbf{z}}_t^c$ follows a zero-mean isotropic Gaussian distribution, the standard deviation of ${\ell}_2$-normalized $\hat{\mathbf{z}}_t^c$ is expected to be approximately $1/\sqrt{d}$, where $d$ represents the dimension of $\hat{\mathbf{z}}_t^c$ along the channel axis. In our case, $d = 16$. A noticeable degeneration of its std from this value suggests a degree of representation collapse.}
\vspace{-0.6cm}
\label{fig_5}
\end{figure}

To quantitatively assess the performance, we compared our method against the following methods:

\begin{itemize}
\item{\textbf{Ours w/o C.L}: This method employs solely the reconstruction approach to directly learn a vision-guided policy without contrastive learning.}
\item{\textbf{Ours w/o Recon}: This method relies only on contrastive learning to obtain an indirect terrain representation without reconstructing the front-facing cube map $\hat{\mathbf{c}}_t^{front}$. }
\item{\textbf{Ours w/o C.A}: This method uses only the self-attention module in the surroundings encoder without the cross-attention module.}
\item{\textbf{Baseline}: This method trains a policy optimized solely by PPO, without any adaptation mechanism.}
\end{itemize}

\subsection{Simulation Experiments}

As shown in Table~\ref{table:1}, we evaluated six different skills under various commands in simulation. We conducted the experiments using 1,000 robots on terrains that were evenly distributed in difficulty, ranging from the simplest to the most challenging. The success rates across these terrains were used as the performance metric. The tested motion skills included jumps with heights of up to 0.7m and distances up to 0.9m, stair climbing with step heights up to 0.15m, traversing overhanging obstacles as low as 0.2m, and camera offset on the z-axis of up to 0.2m. The “blind” experiments simulated scenarios where the RealSense was fully obstructed, introducing depth noise similar to what might occur in deployment when objects are very close to the camera. The z-axis camera offset simulated conditions like moving through tall grass or leaf piles. Our method outperforms others across nearly all tasks. These results suggest that reconstruction provides more direct forward-view information during forward locomotion, enhancing the training process. Additionally, contrastive learning helps the robot extract shared intrinsic features between its own perception and the cube map when moving in other directions, preventing representation collapse caused by excessive discrepancies during supervised learning of the cube map reconstruction. Consequently, the robot exhibits enhanced robustness, even when encountering obstacles beyond its immediate field of view.

\begin{figure}[t]
\centering
\vspace{-0.5cm}
\includegraphics[width=\columnwidth]{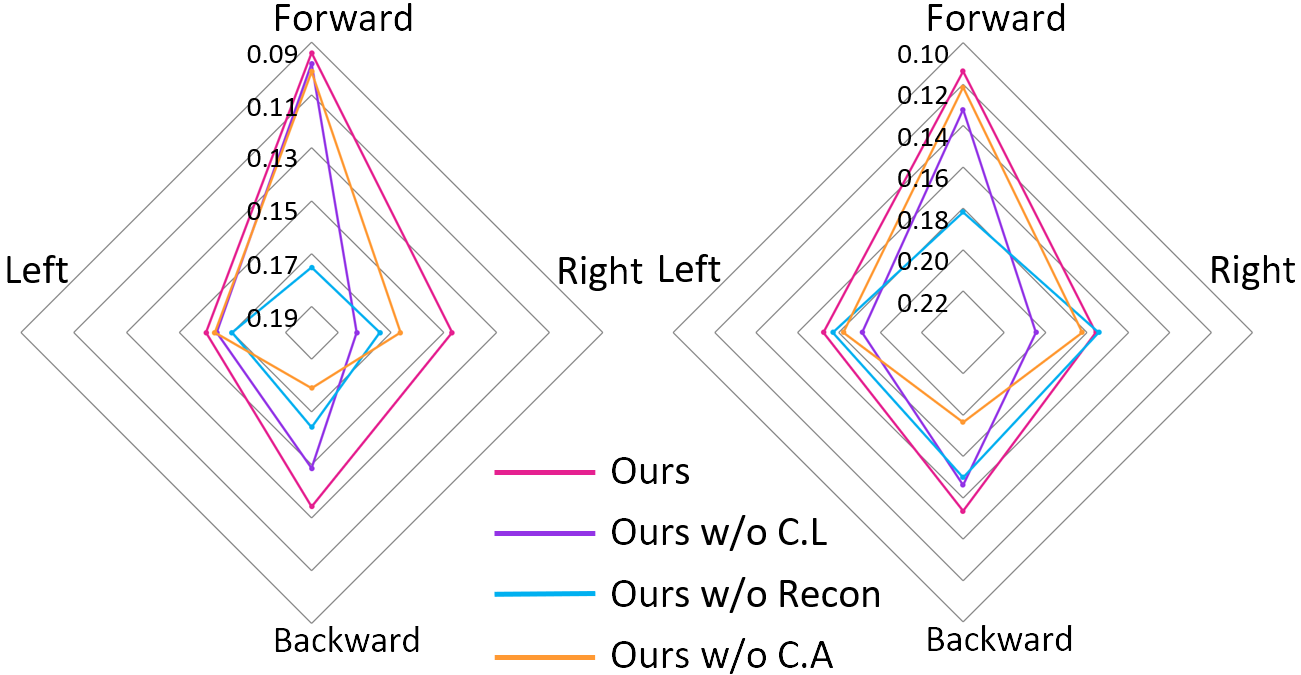}
\vspace{-0.5cm}
\captionsetup{font=footnotesize}
\caption{Average linear velocity tracking error in four directions on 3D complex terrain. \textbf{Left plot:} tested under optimal visual conditions. \textbf{Right plot:} tested with significant visual disturbances.}
\vspace{-0.6cm}
\label{fig_6}
\end{figure}

\begin{figure*}[t]
\centering
\includegraphics[width=7in]{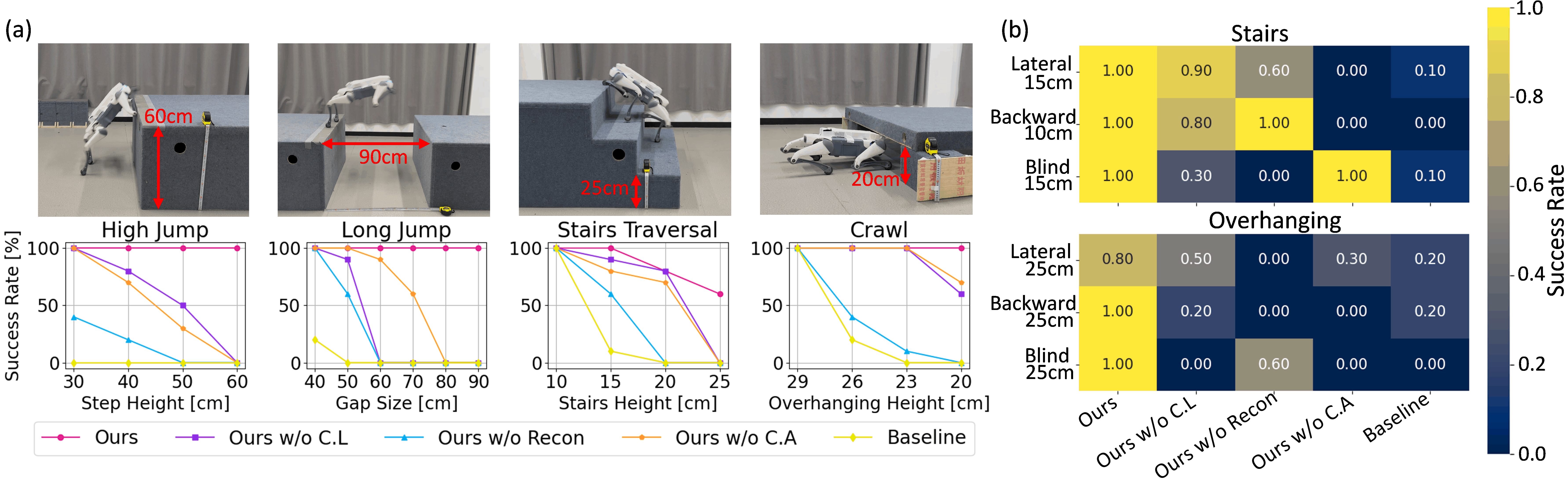}
\vspace{-0.5cm}
\captionsetup{font=footnotesize}
\caption{Real-world indoor quantitative experiments. We evaluated the success rates of our method and the ablations across different environment sets. (a) Tests performed under good visual conditions during forward locomotion. We measured the success rate on four terrains of varying difficulty to assess the maximum locomotion capabilities of each method. (b) Tests conducted under restricted visual conditions. We evaluated three challenging scenarios where visual input was limited: lateral movement, backward movement, and fully obstructed vision (blind), using stairs and overhanging obstacles as test terrains. The x-axis represents different methods, while the y-axis corresponds to the obstacle height and condition for each scenario.}
\vspace{-0.5cm}
\label{fig_7}
\end{figure*}

\begin{figure}[t]
\centering
\includegraphics[width=\columnwidth]{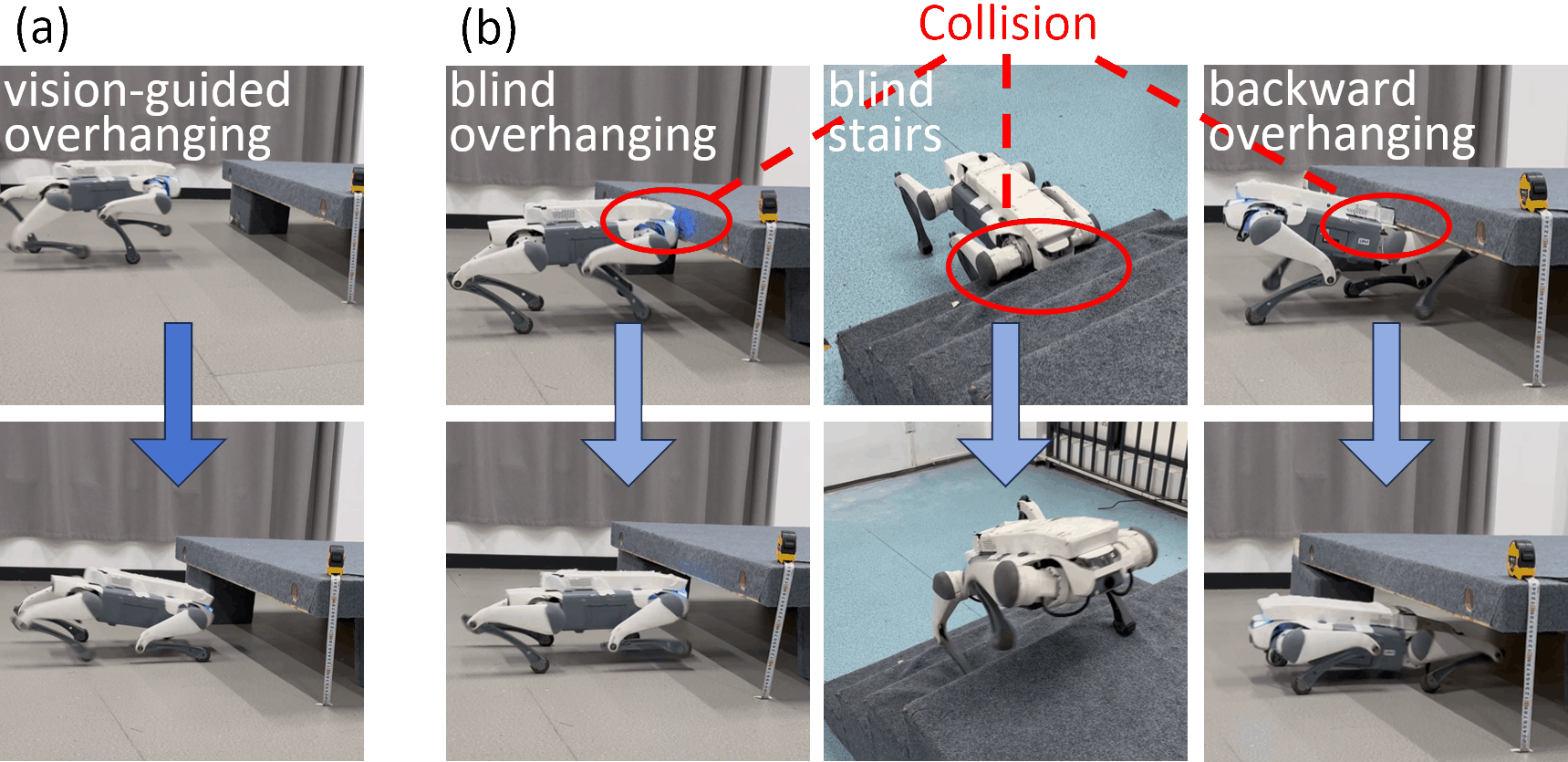}
\vspace{-0.5cm}
\captionsetup{font=footnotesize}
\caption{Qualitative experiments on traversability under different conditions. The first row shows the robot encountering a specific terrain, while the second row illustrates how the robot successfully navigates through it. (a) Traversing overhanging obstacles with a clear line of sight: the robot crouches upon detecting the obstacle, avoiding a collision. (b) Overcoming obstacles in three scenarios where vision is ineffective: the robot relies on proprioceptive feedback to respond to the obstacle, finding a way through after making contact.}
\vspace{-0.5cm}
\label{fig_8}
\end{figure}

To illustrate the effectiveness of our proposed asymmetric attention mechanism from a representation learning perspective, we present the average per-channel standard deviation of the ${\ell}_2$-normalized $\hat{\mathbf{z}}_t^c$, as shown in Fig.~\ref{fig_5}, to quantify the degree of collapse in contrastive learning. Ours exhibits a significantly higher and smoother curve, indicating that our asymmetric attention mechanism successfully mitigates the collapse in contrastive learning solutions, resulting in enhanced representation capability of the surroundings.

Since our primary control objective focus on regulating the robot's speed, Fig.~\ref{fig_6} illustrates the omnidirectional speed tracking performance across complex terrains with both overhanging and underlying obstacles. Ours demonstrates better performance compared to others in both blind and non-blind scenarios, which shows that our method achieves effective omnidirectional speed tracking even in 3D environments with incomplete and noisy vision input.

\subsection{Real-World Experiments}

Extensive real-world experiments were conducted to validate our method's zero-shot sim-to-real capability. As shown in Fig.~\ref{fig_7}, our method consistently outperforms others across various terrains and movement directions. In forward motion, our framework can enable the robot to climb a 0.6m step, leap across a 0.9m gap, ascend 0.25m stairs, and crawl through 0.2m holes. In omnidirectional movement, it can climb 0.15m stairs and pass through 0.25m holes, and it is capable of responding promptly to disturbances in vision through proprioceptive feedback as shown in Fig.~\ref{fig_8}. As far as we know, no existing robot equipped with an egocentric camera has demonstrated all these capabilities simultaneously.

\begin{figure}[t]
\centering
\includegraphics[width=\columnwidth]{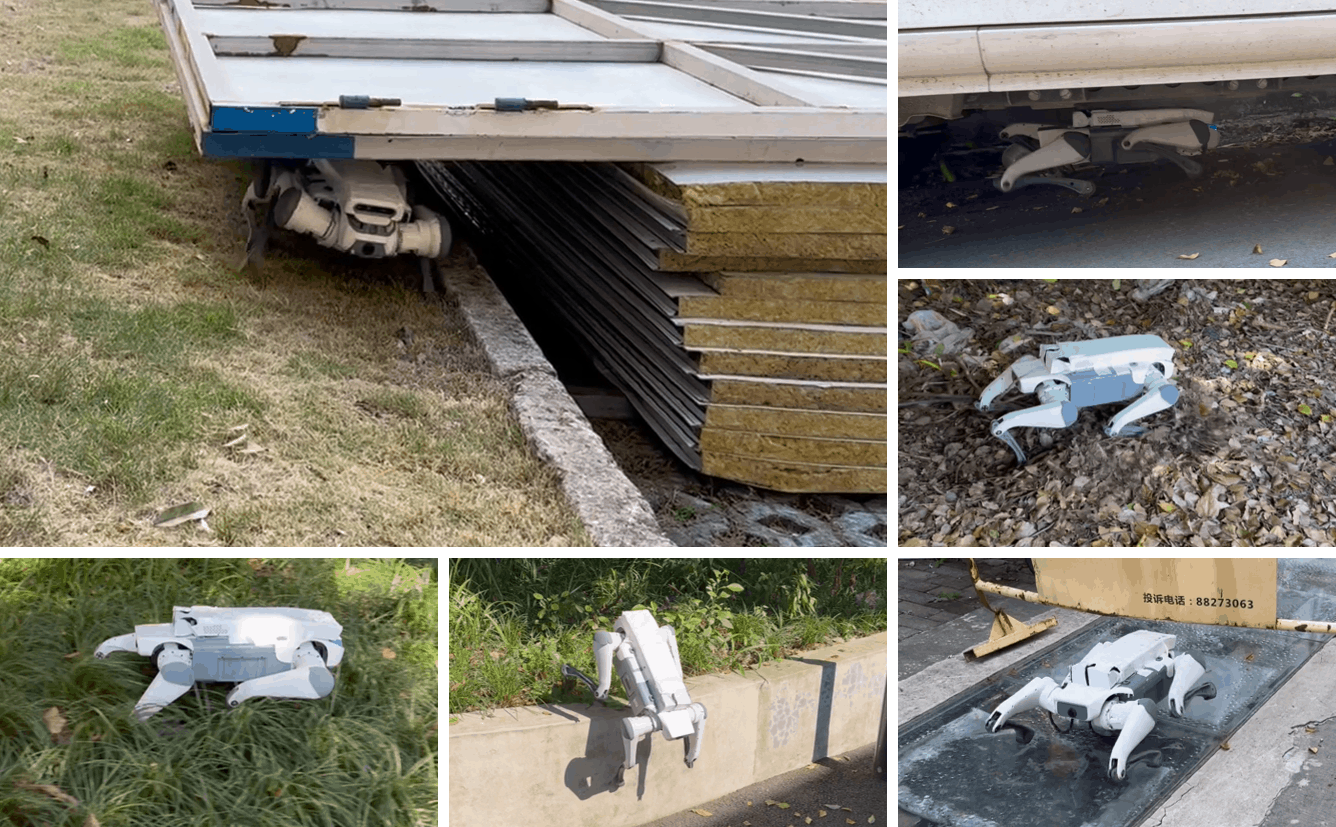}
\vspace{-0.5cm}
\captionsetup{font=footnotesize}
\caption{Outdoor experiments in the wild. We conducted extensive experiments in a variety of complex environments, including wooded areas, roads with parked cars, and construction sites. In scenarios with clear visual perception, the robot could easily jump onto high platforms and crawl omnidirectionally under vehicles. Even when the RealSense was obstructed—such as in tall grass, piles of leaves, or on reflective glass floors—the robot was still able to interact with the environment effectively. It lifted its leg when tripped by grass or twigs, and lowered its head when struck, demonstrating adaptive responses to physical obstacles.}
\vspace{-0.5cm}
\label{fig_9}
\end{figure}

Additionally, as illustrated in Fig.~\ref{fig_9}, we performed extensive experiments in the wild, including omnidirectional mobility over 3D terrains, handling complex environmental noise and disturbance, demonstrating extreme parkour abilities, and integrating all these scenarios. Our method exhibits exceptional robustness, detailed in the supplementary video.

\section{CONCLUSION}
\label{chap:6}

In this work, we propose a novel end-to-end learning framework MOVE, which enables multi-skill omnidirectional legged locomotion with limited view in 3D environments. By incorporating PS-Net with mixed learning method, the robot can  perform omnidirectional movement on 3D terrains without being affected by incomplete visual information. During forward motion, it can effectively extract visual features to execute extreme behaviors such as climbing and jumping. Our results expand the applicability of quadruped robots with egocentric vision in more complex environments.

However, certain limitation still exists. Our exteroception relies solely on depth images, lacking richer semantic information provided by RGB images. In the future, we aim to design a new sensorimotor integration framework, which obtains abundant semantic information from RGB images to achieve better adaptability.

\newpage


\bibliographystyle{IEEEtran}

\bibliography{references}

\end{document}